\documentclass[11pt]{article}
\usepackage{times}
\begin{document}

\begin{center}
{\LARGE The Book of Why: Review}
\end{center}

\begin{center}
{\Large Joseph Y. Halpern}
\end{center}

\bigskip

Just about everyone knows that correlation is not causation, but what
exactly is causation?  Judea Pearl has spent over two decades
trying to understand causation, to define it, and to develop
techniques for inferring it.  This work is having a great impact, and
will arguably ultimately have as great an impact as Pearl's earlier
work on Bayesian networks.

Pearl's landmark book \emph{Causality} was a technical introduction to
his work on the topic.  \emph{The Book of Why} is meant to be a
more popular introduction to the work, as well as documenting some of Pearl's
personal journey through causation.  Pearl and his coauthor Dana
Mackenzie are at particular pains to criticize what seems to be
the predominant view in statistics (and in much of modern-day machine
learning): all the information you need is in the the data; if you
have enough data (and enough computing power) you can figure out
anything you might be interested in.

Pearl argues that, in addition to data, you typically need a causal
model to help you understand the data and draw inferences from it.
The model itself is best represented as a graph, where nodes are
labeled by variables and there is an edge from $X$ to $Y$ if $X$
can directly affect $Y$ (more precisely, if there is a setting of the
variables other than $X$ and $Y$ such that changing the value of $X$ results in
a change in the value of $Y$).

\begin{figure}[ht]
\setlength{\unitlength}{.25in}
\begin{center}
\begin{picture}(8,4)
\put(0,0){\circle*{.2}}
\put(8,0){\circle*{.2}}
\put(4,3){\circle*{.2}}
\put(4,3){\vector(-4,-3){4}}
\put(4,3){\vector(4,-3){4}}
\put(-0.7,0.3){$X$}
\put(8.23,0.3){$Y$}
\put(4.2,3.3){$Z$}
\end{picture}
\end{center}
\caption{$Z$ is a common cause of $X$ and $Y$.}\label{fig:confound1}
\end{figure}
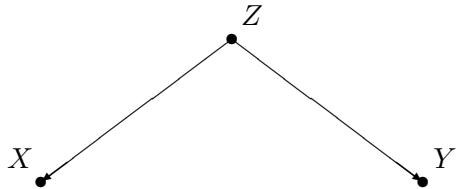

To understand how a causal model can help, let's go back to the
question of correlation vs.~causation.  One reason that $X$ and $Y$
could be correlated is that they have a common cause.  For example,
the causal model in Figure~\ref{fig:confound1} represents a situation
where  $X$ and $Y$ are correlated because they have a common cause
$Z$, but $X$ has no causal impact on $Y$.  

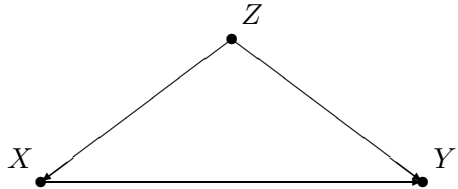
\begin{figure}[ht]
\setlength{\unitlength}{.25in}
\begin{center}
\begin{picture}(8,4)
\put(0,0){\circle*{.2}}
\put(8,0){\circle*{.2}}
\put(4,3){\circle*{.2}}
\put(4,3){\vector(-4,-3){4}}
\put(4,3){\vector(4,-3){4}}
\put(0,0){\vector(1,0){8}}
\put(-0.7,0.3){$X$}
\put(8.23,0.3){$Y$}
\put(4.2,3.3){$Z$}
\end{picture}
\end{center}
\caption{$X$ and $Z$ are both causes of $Y$.}\label{fig:confound2}
\end{figure}

We want to distinguish this situation from the one in
Figure~\ref{fig:confound2}, where 
$X$ and $Y$ still have a common cause but, in addition, $X$ does have
a causal impact on $Y$.  In this case, $Z$ is said to be
a \emph{confounder} (of the causal relationship between $X$ and $Y$).

The standard way to disambiguate the two situations using data is to
control for $Z$; 
that is, set the value of $Z$ and see if then changing $X$ results in a
change in $Y$.  We would expect that, for a fixed setting of $Z$,
changing $X$  would have no impact on $Y$ if the causal model of
Figure~\ref{fig:confound1}  correctly describes the world,  but that
there would be an impact if the world were described by the model in
Figure~\ref{fig:confound2}.

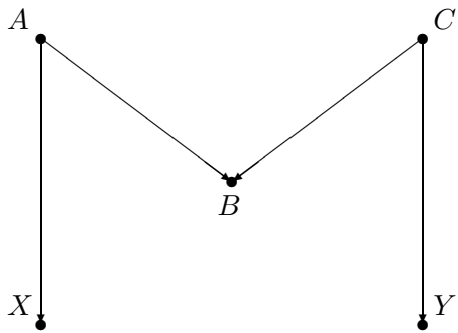
\begin{figure}[ht]
\setlength{\unitlength}{.25in}
\begin{center}
\begin{picture}(8,7)
\put(0,0){\circle*{.2}}
\put(8,0){\circle*{.2}}
\put(8,6){\circle*{.2}}
\put(0,6){\circle*{.2}}
\put(4,3){\circle*{.2}}
\put(0,6){\vector(0,-1){6}}
\put(8,6){\vector(0,-1){6}}
\put(0,6){\vector(4,-3){4}}
\put(8,6){\vector(-4,-3){4}}
\put(-0.7,0.2){$X$}
\put(8.23,0.2){$Y$}
\put(3.7,2.3){$B$}
\put(-0.7,6.2){$A$}
\put(8.23,6.2){$C$}
\end{picture}
\end{center}
\caption{Controlling for $B$ creates a causal connection betwee $X$
  and $Y$.}\label{fig:confound3} 
\end{figure}

It seems that by appropriately controlling for variables (perhaps
using a randomized control trial), we can get back to a situation where the
data gives us all the information we need.    However, the situation
is not so simple.  Consider the causal model in
Figure~\ref{fig:confound3} (taken from p. 161 in \emph{The Book of Why}).
As Pearl points out, in this case, in this case, $X$ has no causal
impact on $Y$, but if we control for $B$, we will ``discover'' a
causal impact.  Controlling for $B$ is the wrong thing to do in this
case.  It creates a causal connection!

The point here is that data by itself may not be enough.   But that
leads to an obvious question: When can we determine causal
relationships from data?  Over the years, Pearl and his collaborators
have developed a number 
of criteria that allow us to do so \emph{given a causal model}.
Perhaps the best-known criterion is the \emph{back-door criterion},
discussed in Chapter 4.  Here it is:  A set $C$ of nodes in a causal
network satisfies the back-door criterion with respect to nodes $X$
and $Y$ if (1) no element of $C$ is a descendant 
of $X$ or $B$ and  (2) every path from $X$ to $Y$ with an arrow into $X$
is blocked by (an element of) $C$. If the backdoor criterion holds,
then
\begin{equation}\label{eq1}
\Pr(Y=b \mid do(X=a)) = \sum_c \Pr(Y=b \mid X=a, C=c)\Pr(C=c).
\end{equation}
That is, we can compute the effect of intervening on $X$ and setting
it to $a$ (represented by $do(X=a)$) by looking at
the conditional probabilities in the observational data.

The back-door criterion is an example of the power of (the graphical
representation of) causal models.  Pearl does an excellent job of
bringing this out in \emph{The Book of Why}.   However,
while I found the discussion of the back-door
criterion really interesting, I found it a little frustrating not to
have a clear statement of it in Chapter 4.  (The discussion in Chapter
4 did prompt me to look up a clear statement on the web.)

This is an example of one problem I had with the book.  When trying to
write a book for the general public, an author needs to walk a fine
line between giving enough technical detail for the reader to be able
to make sense of technical notions while at the same time not
overwhelming the reader with mathematics and mathematical notation.
Of course, what counts as ``enough'' will depend on the reader.
Some readers will no doubt think that there is already far too much
mathematics in the book.  Nevertheless, 
I think that writing down equation (\ref{eq1}) and walking the reader
slowly through what it means, using the (wonderful) examples already in the
book, would, I believe, have been better for most (or, at least, for many)
readers.

Interestingly, (\ref{eq1}) does appear in Chapter 7, where the
back-door criterion is discussed again, but it comes several pages after 
after an English discussion that, again, I think would have been much
clearer if the equation were introduced beforehand.
Indeed, just the observation that there is a \emph{do}
operator on the left-hand side of the equation, denoting an
intervention, and none of the right-hand side, would already help make
the point that we are trying to calculate the effects of an
intervention from data, and what that means.

As I said, \emph{The Book of Why}, to some extent, documents Pearl's
personal journey through causation.  The first few chapters cover some
of the history of causality, particularly in statistics.  
The heart of the book then focuses on how causal
models can help clarify issues like confounding and the effects of
intervention, with an emphasis on the (seminal) work of Pearl and his
students.  Classical statistics is criticized for not having
developed a meaningful theory of causality, which has led it to
struggle with issues like interventions and confounding, and 
answers to counterfactual questions (what would the probability of $Y=y$
have been if I had set $X$ to 1?). 

Because of the focus on Pearl's work, and particularly on how to compute
causal effects given a causal model, the book does not provide a
comprehensive overview of work on causality (nor does it pretend to).
Still, I would have liked to hear a little more of the work of the
``CMU school'' (particularly that of Peter Spirtes, Clark Glymour, and Richard
Scheines).  I do not know the details of the history, but Pearl says
that Peter 
Spirtes preceded him in model causality graphically, in particular, in
modeling the effects of interventions graphically (p. 244), and he 
credits Spirtes, Glymour, and Scheines ``for their help in pushing me
over the cliff of probabilities into the stormy waters of
causation'' (p. 371).   The CMU group (and others) have focused 
on getting causal graphs from data.  Perhaps more could have been said
about this.

The final word on causality remains to be written.  The ``causal
revolution'' that Pearl refers to is still ongoing (with Pearl very much
still in the vanguard).  \emph{The Book of Why} is a great place for a
non-expert to get a sense of the issues and how causal models have
(finally!) helped put thinking about causality on a firm footing.

\medskip

{\bf Acknowledgments:} Halpern's work was supported in part by
NSF grants IIS-178108 and IIS-1703846, a grant from the Open
Philanthropy Foundation, and ARO grant W911NF-17-1-0592.
\end{document}